\documentclass{article}
\usepackage{cite}
\usepackage{amsmath,amssymb,amsfonts}
\usepackage{algorithmic}
\usepackage{graphicx}
\usepackage{textcomp}
\usepackage{xcolor}
\usepackage{enumitem}
\usepackage{color,soul}
\usepackage{booktabs}
\usepackage{multirow}
\usepackage{makecell}
\usepackage{hhline}
\usepackage{algorithm,algorithmic}
\usepackage{url}
\usepackage[colorlinks]{hyperref}
\usepackage{colortbl}
\usepackage{scalerel}
\usepackage{bigstrut}
\usepackage{newfloat}
\usepackage{listings}
\usepackage{color}
\usepackage{bbding}
\usepackage{etoolbox}
\makeatletter
\patchcmd{\@makecaption}
  {\scshape}
  {}
  {}
  {}
\makeatletter
\patchcmd{\@makecaption}
  {\\}
  {.\ }
  {}
  {}
\makeatother

\newlength\savedwidth

\newlength\savewidth
\newcommand\shline{\noalign{\global\savewidth\arrayrulewidth
                            \global\arrayrulewidth 1.5pt}%
                   \hline
                   \noalign{\global\arrayrulewidth\savewidth}}
\def\BibTeX{{\rm B\kern-.05em{\sc i\kern-.025em b}\kern-.08em
    T\kern-.1667em\lower.7ex\hbox{E}\kern-.125emX}}

\definecolor{background_gray}{gray}{0.9}
\usepackage{ijcai24}
\usepackage{times}
\usepackage{soul}
\usepackage{url}
\usepackage[utf8]{inputenc}
\usepackage[small]{caption}
\usepackage{graphicx}
\usepackage{amsmath}
\usepackage{amsthm}
\usepackage{booktabs}
\usepackage{algorithm}
\usepackage{algorithmic}
\usepackage[switch]{lineno}
\usepackage[colorlinks]{hyperref}

\title{Predicting Parking Availability in Singapore with Cross-Domain Data: \\ A New Dataset and A Data-Driven Approach

}

\author{
Huaiwu Zhang$^{1}$\textsuperscript{*}\and
Yutong Xia$^{2}$\textsuperscript{*}\and
Siru Zhong$^{1}$\and
Kun Wang$^2$\and
Zekun Tong$^2$\and \\
Qingsong Wen$^3$\and
Roger Zimmermann$^2$\And
Yuxuan Liang$^{1,4}$\textsuperscript{†}\\
\affiliations
$^1$The Hong Kong University of Science and Technology (Guangzhou), China \\ 
$^2$National University of Singapore, Singapore \\
$^3$Squirrel AI, USA\\
$^4$State Key Laboratory of Resources and Environmental Information System, China
\emails
\{huaiwuzhang1998, yutong.x, siruzhong, yuxliang\}@outlook.com,
wk520529@mail.ustc.edu.cn,
zekuntong@u.nus.edu,
qingsongedu@gmail.com,
rogerz@comp.nus.edu.sg
}

\begin{document}

\maketitle

\begin{abstract}
The increasing number of vehicles highlights the need for efficient parking space management. Predicting real-time Parking Availability (PA) can help mitigate traffic congestion and the corresponding social problems, which is a pressing issue in densely populated cities like Singapore. In this study, we aim to collectively predict future PA across Singapore with complex factors from various domains. The contributions in this paper are listed as follows: \textit{(1) A New Dataset:} We introduce the \texttt{SINPA} dataset, containing a year's worth of PA data from 1,687 parking lots in Singapore, enriched with various spatial and temporal factors. \textit{(2) A Data-Driven Approach:} We present DeepPA, a novel deep-learning framework, to collectively and efficiently predict future PA across thousands of parking lots. \textit{(3) Extensive Experiments and Deployment:} DeepPA demonstrates a 9.2\% reduction in prediction error for up to 3-hour forecasts compared to existing advanced models. Furthermore, we implement DeepPA in a practical web-based platform to provide real-time PA predictions to aid drivers and inform urban planning for the governors in Singapore. We release the dataset and source code at \url{https://github.com/yoshall/SINPA}.
\end{abstract}

\renewcommand{\thefootnote}{\fnsymbol{footnote}}
\footnotetext[1]{H. Zhang and Y. Xia contributed equally to this work.}
\footnotetext[2]{Corresponding author.  Email: yuxliang@outlook.com}

\section{Introduction}\label{para:intro}

Urban traffic congestion significantly contributes to air pollution and poses health risks~\cite{kelly2015air,liang2017inferring}. The uneven distribution of parking spaces and inadequate parking decision-making by drivers are significant factors contributing to traffic congestion~\cite{tilahun2017cooperative}. The challenge of finding parking spaces is expected to grow as the population increases, thus the resulting social and environmental problems will be especially more acute in densely populated countries such as Singapore~\cite{shamsuzzoha2021smart}. To address these problems, there is an urgent need to optimize this issue by employing precise and efficient prediction of \textbf{Parking Availability (PA)} using smart city technologies~\cite{xie2022maas}. 

\begin{figure}[!t]
  \centering
   \includegraphics[width=\linewidth]{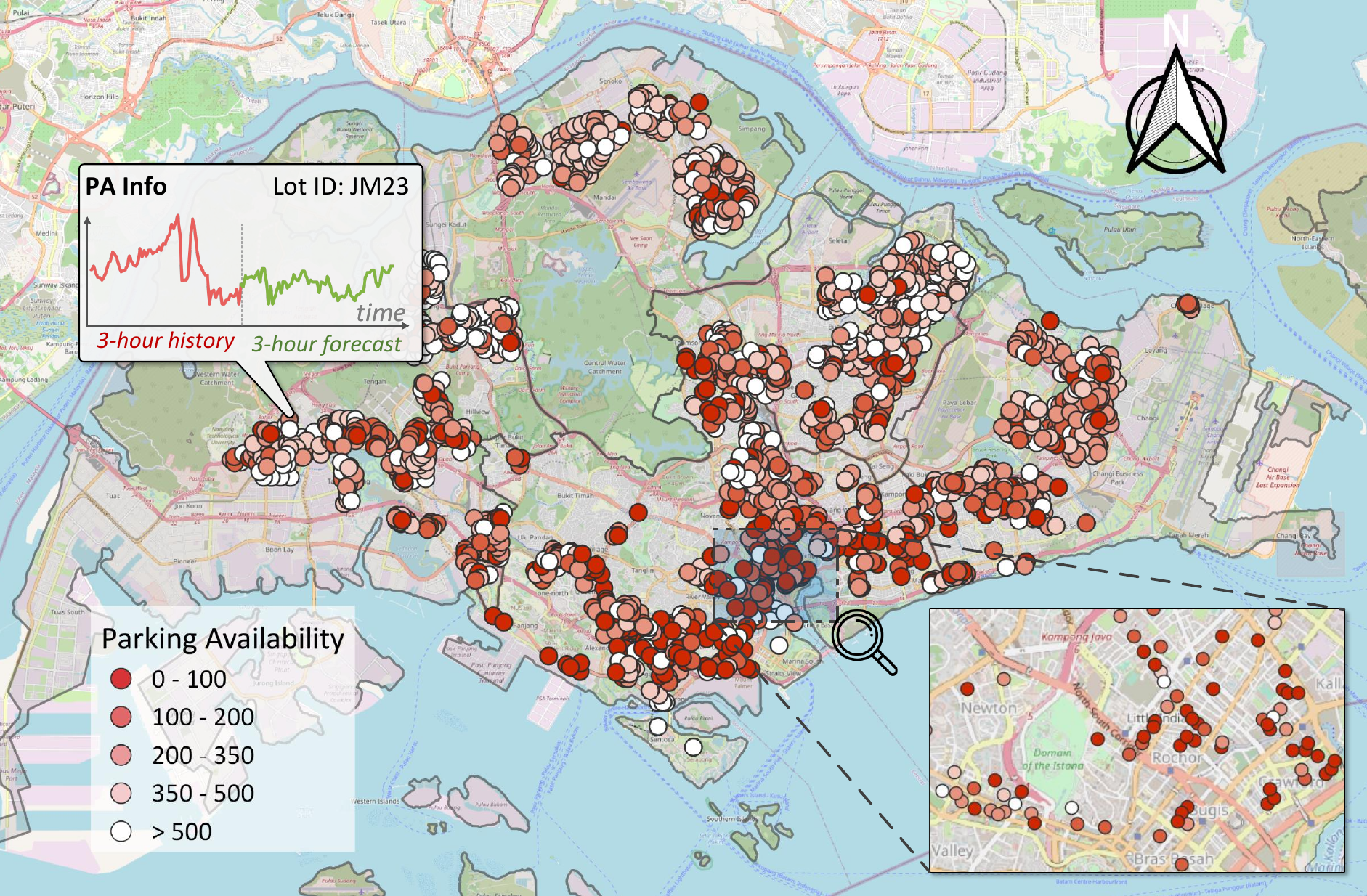}
  \caption{Distribution of 1,687 parking lots throughout Singapore. Each lot reports real-time parking availability  every 15 minutes.}\label{fig:intro}
\end{figure}

In this study, we focus on collectively predicting PA across thousands of parking lots in Singapore, as shown in Figure~\ref{fig:intro}. This undertaking is confronted with several noteworthy challenges. Firstly, the availability of parking spaces within a parking lot is intricately influenced by multiple \textit{complex factors}, including both temporal and spatial dimensions. Temporal external factors (e.g., weather conditions) wield substantial influence in shaping travel preferences and, consequently, parking patterns. Meanwhile, spatial factors, such as land use and road density, exert a noteworthy impact on the parking behavior of residents, consequently affecting PA. Therefore, effectively capturing the effects of temporal and spatial factors on PA is a necessary and formidable challenge.

Secondly, the \emph{complex spatial correlations} among parking lots exhibit a non-linear relationship concerning geospatial distance measured in Euclidean space. This non-linearity is further compounded by the inherent disparities in their attributes, such as land functions and road network structures. In essence, the PA patterns of two proximate parking lots may diverge significantly due to distinctions in their intrinsic spatial characteristics. Additionally, comprehending the interconnected dynamics among various parking lots, influenced by the relatively stable total number of vehicles in the region, is imperative for a better understanding of PA patterns.

Thirdly, an approach that captures long-range (i.e., global) spatial dependencies \cite{zhang2022design} comes with another challenge in terms of \emph{computational efficiency}. This challenge is particularly pronounced at fine spatial granularity, i.e., thousands of carparks in our task. Common methods like Graph Convolutional Networks\cite{kipfsemi} capture global spatial dependencies by stacking many local convolutional layers, which significantly increases computational costs. More advanced methods \cite{wu2019graph,wu2020connecting} using an adaptive adjacency matrix also result in a high computational complexity i.e., $\mathcal{O}(N^2)$, and the computational load of methods like Multi-head Self-Attention (MSA)\cite{vaswani2017attention} also increases quadratically with the number of nodes $N$. Therefore, efficiently modeling the global spatial dependencies remains an unresolved issue.

Given the challenges outlined above, we aim to develop a model that not only incorporates a variety of cross-domain factors but also effectively captures the complex spatial correlations among parking lots to predict future PA accurately. To achieve this, we first crawl and process the \textbf{Sin}gapore \textbf{PA} dataset, i.e., \texttt{SINPA}, including PA data along with various external spatio-temporal factors, such as meteorological and land use data, specific to Singapore. 

We then propose a data-driven model \textbf{DeepPA}, for collectively forecasting the future PA readings across thousands of carparks in Singapore while maintaining a manageable computational load. Specifically, we introduce the Graph Cosine Operator (GCO), a new mechanism for dynamically capturing spatial dependencies across extensive parking lots while mitigating computational complexity through discrete cosine transform. In addition, considering the crucial role of modeling temporal dependencies \cite{liang2018geoman,guo2019attention,zheng2020gman}, we innovatively treat temporal information as a distinct ``parking lot'' entity. This allows us to integrate spatial data into position encoding effectively. Then we utilize causal MSA \cite{oord2016wavenet,liang2023airformer} to adhere to temporal sequences, thereby facilitating the learning of time-sensitive patterns in a manner that is both efficient and interpretable.

\renewcommand{\thefootnote}{\arabic{footnote}}
\setcounter{footnote}{0} 

In summary, our contributions lie in the following aspects:
\begin{itemize}[leftmargin=*]
    \item \textbf{A New Dataset}. We crawl, process, and introduce \texttt{SINPA}, a large-scale parking availability dataset incorporating cross-domain data in Singapore. To the best of our knowledge, this dataset is the first publicly available dataset in the field of PA values forecasting, providing diverse applications within spatio-temporal domain research. The dataset is publicly accessible at \url{https://github.com/yoshall/SINPA}. It includes data obtained from Data.gov.sg, Urban Redevelopment Authority, and Land Transport Authority, which are made available under the Singapore Open Data Licence\footnote{\url{https://beta.data.gov.sg/open-data-license}.}

    \item \textbf{A Data-Driven Approach}. Leveraging insights gained from the analysis of multi-domain features, we develop a simple yet effective deep learning framework, DeepPA, to predict PA readings. 
    DeepPA uses Graph Cosine Operator and Casual MSA to capture the complex spatial and temporal correlations among parking lots while ensuring an acceptable computational load.

    \item \textbf{Extensive Experiments}. We evaluate DeepPA on our dataset. The empirical results demonstrate the model's high accuracy, efficiency, and adaptability. These attributes make DeepPA an ideal candidate for real-world deployment for real-time PA prediction applications.

\end{itemize}

\section{Preliminary}\label{sec:pre}

Parking Availability (PA) refers to the remaining space for parking in a parking lot. At the time $t$, the PA of the parking lots can be denoted as $\mathbf{X}^{t} = \left\{x^{t}_{1}, x^{t}_{2}, ..., x^{t}_{N}\right\} \in \mathbb{R}^{N}$, where $N$ refers to the total number of parking lots. Each entry $x^{t}_{n}$ indicates the count of remaining parking spaces of $n$-th parking lots at time $t$. Given the historical PA of all carparks from the past $T$ time steps and corresponding external features, we aim to learn a function $f(\cdot)$ that predicts their PA readings over the next $\tau$ steps: 
\begin{equation}
    f(\mathbf{X}^{1:T}; \mathbf{F}_{t}^{1:T}, \mathbf{F}_{s}) \rightarrow \mathbf{Y}^{1:\tau},
\end{equation}
where $\mathbf{X}^{1:T} \in \mathbb{R}^{T \times N }$ indicates historical PA of all carparks; $\mathbf{F}^{1:T}_{t} = \left\{F ^{1:T}_{mete}, F^{1:T}_{time} \right\} \in \mathbb{R}^{T \times C_{t}}$ denotes external temporal features and $C_{t}$ represent its dimension; $\mathbf{F}_{s} = \left\{F_{loc}, F_{pln}, F_{use}, F_{rd} \right\} \in \mathbb{R}^{N \times C_{s}}$ denotes the spatially related feature and $C_{s}$ represent its dimension; $\mathbf{Y}^{1:\tau} \in \mathbb{R}^{\tau \times N}$ is the future PAs. 

\subsection{Related Work}
\subsubsection{Intelligent Parking}

Intelligent parking has attracted considerable attention from both academia and industry. Regarding the prediction of PA, Sii-MOBILITY\cite{badii2018predicting} integrates weather and traffic flow predicting PA based on Bayesian regularization neural networks. Subsequently, \cite{zhang2020semi} proposes a novel method SHARE based on Gated Recurrent Units\cite{chung2014empirical} and contextual graph convolution to capture both local and global spatial dependencies within parking lots, enabling estimation of missing PA in both temporal and spatial dimensions. 

In addition, intelligent parking systems have other downstream tasks. \cite{duparking} employs Long Short-Term Memory (LSTM) to model the PA-related temporal closeness, period, and current general influence to achieve parking availability imputation. Utilizing PA prediction combined with recommendation algorithms suggested the optimal parking lots near the user's specified destination \cite{xu2023empowering}. Moreover, models such as Random Forest and LSTM have identified the influence of weather factors on drivers' parking behavior \cite{zhang2020pewlstm}. Besides, unlike enclosed parking lots that are independent of the street, it has also achieved the prediction of the availability of curbside parking spaces \cite{roman2018detecting}. 

\begin{figure*}[!t]
  \centering
  \includegraphics[width=0.93\linewidth]{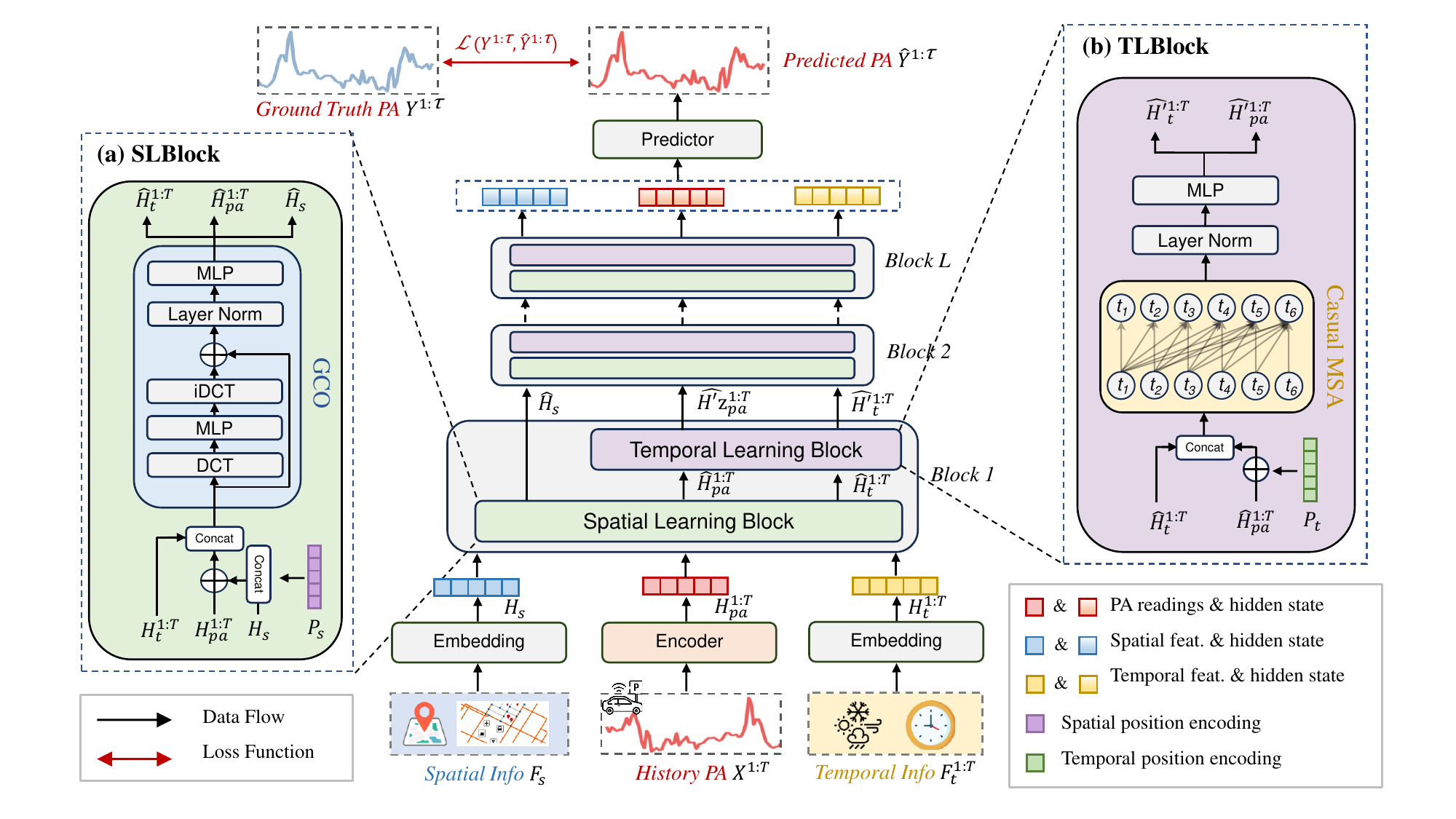}
  \caption{The framework of our proposed DeepPA. Initially, historical PA readings and spatio-temporal information are encoded and mapped into hidden states, respectively. Subsequently, PA will interact with spatio-temporal information in two sequential stages: (a) Spatial Learning Block (SLBlock) captures spatial dependence among parking lots, in which we introduce a novel structure Graph Cosine Operator (GCO) to enhance interaction efficiency; (b) TLBlock simultaneously investigates the temporal dynamics of PA and temporal information adhering to temporal patterns through Causal Multi-Head Attention (Causal MSA). Finally, DeepPA will integrate the feature space to forecast future PA. Concat: Concatenate. Info:Information }
  \label{fig:model}
\end{figure*}
\subsubsection{Data-Driven Spatio-Temporal Forecasting}
Spatio-Temporal (ST) forecasting has garnered significant attention in the past few years~\cite{jin2023spatio}. Conventional models, e.g., ARIMA and VAR, have enjoyed widespread adoption, while their performance tends to be suboptimal with highly volatile datasets\cite{liang2018geoman,guo2019attention,zheng2020gman,liang2021revisiting}. Building upon Graph Neural Networks (GNN) \cite{kipfsemi}, Spatio-Temporal Graph Neural Networks (STGNN) \cite{wang2021survey} have emerged for ST data by integrating temporal components, e.g. Temporal Convolutional Networks \cite{BaiTCN2018} or Recurrent Neural Networks \cite{graves2013generating}. Pioneering examples include DCRNN \cite{li2018dcrnn_traffic}, STGCN \cite{yu2018spatio}, and ST-MGCN \cite{geng2019spatiotemporal}. Subsequent studies GWNet \cite{wu2019graph} and AGCRN \cite{bai2020adaptive}, leverage adaptive adjacency matrices to enhance predictive performance. Meanwhile, ASTGCN \cite{guo2019attention}, GMAN \cite{zheng2020gman}, and STTN\cite{xu2020spatial} employ attention mechanisms to discern dynamic ST dependencies. STGODE \cite{fang2021stgode}, STGNCDE \cite{stgncde}, and MixRNN~\cite{liang2022mixed} capture continuous ST dynamics by employing neural ordinary differential equations. Causality-based methods such as CaST \cite{xia2024deciphering} resolve the distribution shift problem via causal tools. Nonetheless, these methods are not tailored to the parking problem while concurrently addressing the above-mentioned challenges. 

\section{Methodology}\label{sec:method}

The spatial and temporal external factors (i.e., weather and geo-location) play an important role in PA forecasting. To address the challenges of complex spatial correlations and computational efficiency, we have developed DeepPA, a deep learning framework for efficiently predicting future PA throughout Singapore. 
\subsection{Overall Framework}
To address the challenges of complex spatial correlations and computational efficiency, we develop DeepPA, a deep learning framework for efficiently predicting future PA throughout Singapore, shown in Figure \ref{fig:model}.
We first transform the historical PA $\mathbf{X}^{1:T}$, along with spatial features $\mathbf{F}_{s}$ and temporal features $\mathbf{F}_{t}^{1:T}$ into the latent feature space using encoders, implemented through Multi-Layer Perceptrons (MLPs) and embedding layers. The hidden features are then fed to $L$ DeepPA blocks to interact with each other to capture ST dependencies within data. Each DeepPA block consists of two sub-blocks as follows.
\begin{itemize}[leftmargin=*]
    \item \emph{Spatial Learning Block (SLBlock)}: This module is utilized to capture intricate spatial relationships between parking lots, adopting a unique perspective by treating time information as a new node. To efficiently capture the non-Euclidean spatial relationship between thousands of parking lots, we introduce a novel Graph Cosine Operator.

    \item \emph{Temporal Learning Block (TLBlock)}: This transformer-structure module is dedicated to capturing the periodic patterns that emerge over time at each parking lot. Specifically, it simultaneously model the hidden state of PA $\mathbf{H}^{1:T}$ and temporal features $\mathbf{F}_{t}^{1:T}$. To preserve the sequential order, we incorporate a mask in the MSA mechanism, ensuring that each state focuses only on its preceding states. 
\end{itemize}
 
\subsection{Spatial Learning Block (SLBlock)}
Considering PA across different parking lots is interdependent, we aim to develop a module that effectively captures the specific spatial dependencies among thousands of parking lots while ensuring lower computational costs.

\subsubsection{Overview}

As depicted in the left part of Figure \ref{fig:model}, $\mathbf{H}_{s} \in \mathbb{R}^{B \times T \times N \times C_{hs}}$ is concatenated with the spatial position encoding $\mathbf{P_{s}} \in \mathbb{R}^{B \times T \times N \times (C_{p}-C_{hs})}$ and then added to $\mathbf{H}^{1:T}_{pa}$:
\begin{equation}
    \mathbf{H}^{1:T} = \mathbf{H}^{1:T}_{pa} + \operatorname{concat}(\mathbf{H}_{s}, \mathbf{P_{s}}), \mathbf{H}^{1:T} \in \mathbb{R}^{B \times T \times N \times C_{p}}
\end{equation}
where $\mathbf{H}^{1:T}$ is the PA with integrated the spatial features. In order to achieve interaction with temporal information, we then map the temporal information to the same dimension $C_{p}$ as $\mathbf{H}^{1:T}$, treating it as an auxiliary node, i.e., a virtual node which acts as the $(N+1)$-th parking lot, and concatenating it with $\mathbf{H}^{1:T}$:
\begin{equation}
    \mathbf{H}^{1:T} = \operatorname{concat}(\mathbf{H}^{1:T}, \mathbf{H}^{1:T}_{t}), \mathbf{H}^{1:T} \in \mathbb{R}^{BT \times (N+1) \times C_{p}}
\end{equation}
where $\mathbf{H}^{1:T}$ represents the PA that integrates spatio-temporal information. After obtaining the initial features from both spatial and temporal domains, we jointly capture local and global spatial correlations of $\mathbf{H}^{1:T}$ via our proposed \emph{Graph Cosine Operator (GCO)} by $\mathbf{\hat{H}}^{1:T} = \operatorname{GCO}(\mathbf{H}^{1:T})$, which will be described in the subsequent part.

\subsubsection{Graph Cosine Operator (GCO)}
Since dynamic adjacency matrix or attention mechanism in our dataset is computationally extensive, we design GCO to mitigate this issue by simulating how spatial dependencies induce changes $\Delta\mathbf{\hat{H}}^{1:T}$ in the hidden state of the parking lot. For a single parking lot, the variation of PA at each moment is known to be affected by all other parking lots, which can be expressed by the \emph{heat conduction formula} \cite{wang2007heat}:
\begin{equation}
    \Delta\hat{H}^{t}_{i} = \sum k_{ji}(\hat{H}^{t}_{j}-\hat{H}^{t}_{i})\ \ (i,j \in N\ , t \in T)
\end{equation}
where $k_{ji}$ represents the thermodynamic coefficient, which, in our study, signifies a learnable correlation coefficient between ${\hat{H}}^{t}_{j}$ and ${\hat{H}}^{t}_{i}$. Thus, at the same time instant $t$, the changes in the hidden states of all parking lots, i.e., $\Delta\mathbf{\hat{H}}^{t}$, can be represented as follows.
\begin{equation} 
    \begin{aligned}
    \Delta\mathbf{\hat{H}}^{t} = 
    \left[
    \begin{array}{c}
         \Delta\hat{H}^{t}_{1}\\
         \Delta\hat{H}^{t}_{2}\\
         ...\\
         \Delta\hat{H}^{t}_{N}
    \end{array}
    \right]
    =
    \left[
     \begin{array}{c}
         \sum k_{j1}(\hat{H}^{t}_{j}-\hat{H}^{t}_{1})\\
         \sum k_{j2}(\hat{H}^{t}_{j}-\hat{H}^{t}_{2})\\
         ...\\
         \sum k_{jN}(\hat{H}^{t}_{j}-\hat{H}^{t}_{N})
     \end{array}
    \right] \\
    \end{aligned}\label{equ:deltaxt}
\end{equation}

Since the number of vehicles remains essentially constant throughout the entire range, we can assume that the correlation between two parking lots is consistent, i.e., $k_{ji}=k_{ij}$. After applying the distributive property to Eq. \ref{equ:deltaxt}, we can obtain the following expression:
\begin{equation} 
\small
    \begin{aligned}
    \Delta\mathbf{\hat{H}}^{t} =
    \left[
     \begin{array}{ccc}
         k_{11} & \cdots & k_{1N}\\
         \vdots & \ddots & \vdots\\
         k_{N1} & \cdots & k_{NN}
     \end{array}
    \right]
    \mathbf{\hat{H}}^{t}- 
    \left[
     \begin{array}{cccc}
         K_{1} & & &\\
         & K_{2} & &\\
         & &...&\\
         & & & K_{N}
     \end{array}
    \right]
    \mathbf{\hat{H}}^{t} 
    \end{aligned}\label{equ:degree_adjacency}
\end{equation} 
where $K_{i}$ represents $\sum k_{ji}$, $j \in N$. It can be observed that the two learnable matrices in the equation are respectively a symmetric matrix and a diagonal matrix. Hence, they can be regarded as a special form of degree matrix $\mathbf{\hat{D}}$ and adjacency matrix $\mathbf{\hat{A}}$. Therefore, Eq.~\ref{equ:degree_adjacency} can be expressed in the following form:
\begin{equation} 
    \begin{aligned}
    \Delta\mathbf{\hat{H}}^{t} = (\mathbf{\hat{D}}-\mathbf{\hat{A}})\mathbf{\hat{H}}^{t}=\mathbf{\hat{L}}\mathbf{\hat{H}}^{t}=\sigma\mathbf{L}\mathbf{\hat{H}}^{t}
    \end{aligned}\label{equ:laplace}
\end{equation} 
where $\mathbf{L}$ represents the Laplacian matrix \cite{merris1994laplacian}, We aim to fit $\mathbf{\hat{L}}$ by adding a set of trainable weights $\sigma$ to $\mathbf{L}$. Let $\lambda$ and $\mathbf{v}$ be the eigenvalues and eigenvectors of $\mathbf{L}$, respectively. Then Eq.~\ref{equ:laplace} can be expressed as follows: 
\begin{equation} 
    \begin{aligned}
    \mathbf{\hat{L}}\mathbf{\hat{H}}^{t}=\sigma \mathbf{L}\mathbf{v}=\sigma\lambda\mathbf{v}
    \end{aligned}\label{equ:eigenvalues}
\end{equation} 

Left-multiplying by the Laplacian matrix is equivalent to solving for the second derivative \cite{koren2003spectral}. Therefore, for Fourier bases, the transformation can be denoted as follows:
\begin{equation}
    \sigma \mathbf{L} e^{-j\omega t} = \sigma \frac{\partial ^{2} e^{-j\omega t}}{\partial t^{2}} = \sigma(-\omega^{2})e^{-j\omega t}    
\end{equation}
where $-\omega^{2}$ represents the the eigenvalues of the Laplacian matrix. Defining a set of eigenvalues of $\mathbf{L}$ as:
\begin{equation}
    \Lambda=(\lambda_{1},\lambda_{2}, ..., \lambda_{N}) \in (-\omega^{2})
\end{equation}
The corresponding unit eigenvectors are denoted as:
\begin{equation}
    \mathbf{U}=(\vec{u_{1}},\vec{u_{2}}, ..., \vec{u_{N}}) \in e^{-j\omega t}
\end{equation}
Thus Eq.~\ref{equ:eigenvalues} can be rewritten as:
\begin{equation}
    \mathbf{\hat{L}}\mathbf{\hat{H}}^{t} = \sigma \mathbf{U}\Lambda \mathbf{U}^{T}\mathbf{\hat{H}}^{t} =
    \sigma (\mathcal{F}^{-1}(\mathcal{F}(\Lambda)\cdot \mathcal{F}(\mathbf{\hat{H}}^{t}))
\end{equation}
where $\mathcal{F}(\cdot)$ represents Discrete Fourier Transform (DFT). 
Inspired by Fourier Neural Operator (FNO)\cite{li2020fourier}, we parameterize $\sigma$ and $\mathcal{F}(\Lambda)$ via MLP layers. Previous work employed neural operators to approximate the MSA\cite{guibas2021efficient}. According to Euler's formula, the Fourier transform maps data from the time domain to the frequency domain, converting real numbers into complex numbers:
\begin{equation}
    e^{-j\omega t}=\cos{\omega t}+j\sin{\omega t}
\end{equation}
After this transformation, complex numbers are decomposed into their real and imaginary components. When multiplied by weights, it is crucial to adhere to the distributive property of multiplication, thus MLPs exhibit a fourfold increase in computational time when operating on complex numbers compared to real numbers. In the case of retaining only the real part, we use the Discrete Cosine Transform (DCT) to approximate DFT to reduce computational complexity while preserving performance efficacy, as demonstrated in Table~\ref{tab:GCO}. For a clearer comprehension, the overall operational steps of GCO for $\mathbf{H}^{1:T}$ are as illustrated in Algorithm~\ref{algo:gco}.

\begin{algorithm}[!t]
        \renewcommand{\algorithmicrequire}{\textbf{Input:}}
	\renewcommand{\algorithmicensure}{\textbf{Output:}}
 \small
	\caption{Graph Cosine Operator}
    \label{algo:gco}
    \begin{algorithmic}[1] 
        \REQUIRE  $\mathbf{H}^{1:T} \in \mathbb{R}^{BT \times (N+1) \times C_{p}}$; 
	    \ENSURE Updated hidden states $\mathbf{\hat{H}}^{1:T}$; 
        \STATE Step 1: Projection into the frequency domain.
        \\
        $\mathbf{\hat{H}}^{1:T}=\text{DCT}(\mathbf{H}^{1:T}) $
        \STATE Step 2: Fitting the basis functions.
        \\
        $\mathbf{\hat{H}}^{1:T}=\text{MLP}(\mathbf{\hat{H}}^{1:T}) $
        \STATE Step 3: Returning to the spatial domain.
        \\
        $\mathbf{\hat{H}}^{1:T}=\text{iDCT}(\mathbf{\hat{H}}^{1:T}) $
        \STATE Step 4: Normalizing the output.
        \\
        $\mathbf{\hat{H}}^{1:T}=\text{LayerNorm}(\mathbf{\hat{H}}^{1:T}) $
        \STATE Step 5: Fitting the Laplacian matrix.
        \\
        $\mathbf{\hat{H}}^{1:T}=\text{MLP}(\mathbf{\hat{H}}^{1:T}) $
    \end{algorithmic}
\end{algorithm}

\subsubsection{Discussion}
SLBlock is designed by jointly considering the following factors: 1) \emph{Spatial dependencies}. Given that the strength of mutual influence between stations is not solely dependent on distance, we introduced a dynamic correlation coefficient $k_{ji}$ to calculate the relationship between parking lots. 2) \emph{Efficiency}. Via the GCO, we have managed to reduce the computational complexity.
3) \emph{Temporal factor interation}: We treat the overall temporal information as a new node interacting with the PA to take temporal external factors into account.

\subsection{Temporal Learning Block (TLBlock)}
Accurately capturing the periodic relationship is crucial for the prediction of PA. Meanwhile, variations in weather conditions and the impact of holidays can alter the travel patterns of residents, thereby impacting this periodicity. Therefore, we designed a novel module TLBlock to capture the periodicity while taking into account the impact of special time variables.

\subsubsection{Capturing temporal correlation}
First, to enable position-aware MSA, we infuse learnable position encoding into the input PA of TLBlock. 
Then, we decouple $\mathbf{\hat{H}}^{1:T}_{t}$ and $\mathbf{\hat{H}}^{1:T}$, where $\mathbf{\hat{H}}^{1:T}$ is considered as the $(N+1)$th parking lot in the SLBlock:
    $\mathbf{\hat{H}}^{1:T}, \mathbf{\hat{H}}^{1:T}_{t}=\operatorname{split}(\mathbf{\hat{H}}^{1:T})$.
Next, we merge the $B$ and $N$ dimensions of $\mathbf{\hat{H}}^{1:T}$ and concatenate them with $\mathbf{\hat{H}}^{1:T}_{t}$:
\begin{equation}
\nonumber
    \mathbf{\hat{H}}^{1:T} = \operatorname{concat}(\mathbf{\hat{H}}^{1:T}, \mathbf{\hat{H}}^{1:T}_{t}), \mathbf{\hat{H}}^{1:T} + \mathbf{P_{t}} \in \mathbb{R}^{(BN+B) \times T \times C_{p}}
\end{equation}
where $\mathbf{P_{t}}$ represents temporal position encoding. Subsequently, temporal factors are modeled alongside PA using the same encoder to enable the model will take into account the influence of temporal information to a greater extent.

\subsubsection{Temporal causality}

Given that the PA at a given step is not contingent on its future, inspired by \cite{oord2016wavenet,liang2023airformer}, we introduce causality into MSA to ensure the model adheres to the temporal sequence of the input data, implemented by masking specific entries in the attention map:
\begin{equation}
    \begin{aligned}
    \mathbf{Q} = \mathbf{W_{q}}\mathbf{\hat{H}}^{1:T}; \quad
    \mathbf{K} = \mathbf{W_{k}}\mathbf{\hat{H}}^{1:T}; \quad
    \mathbf{V} = \mathbf{W_{v}}\mathbf{\hat{H}}^{1:T}
    \\
    \text{Causal MSA}(\mathbf{\hat{H}}^{1:T}) = \text{softmax}(\frac{\mathbf{Q}\mathbf{K}^T}{\sqrt{\alpha}}+\mathbf{M})\mathbf{V}
    \end{aligned}
\end{equation}
 For a single state, $\hat{H}^{i}$, applying a mask $\mathbf{M} \in \mathbb{R}^{\alpha \times T \times T \times C_{p}/\alpha}$ involves setting its attention weights towards temporally later states $\hat{H}^{j}$ to $-\inf$, where $j \in (t+1, T)$.

\subsection{Prediction \& Optimization}

DeepPA makes predictions based on the hidden state of the last DeepPA block $\mathbf{\hat{H}}^{1:T}$ using a predictor, which is implemented via an MLP layer in our implementation:
\begin{equation}
    \mathbf{\hat{Y}}^{1:\tau} = \operatorname{MLP}(\mathbf{\hat{H}}^{1:T}).
\end{equation}
To train our model, we minimize the loss function $\mathcal{L}_{MAE}$. Here, $\mathcal{L}_{MAE}$ is the Mean Absolute Error for evaluating the errors between our prediction $\mathbf{\hat{Y}}^{1:\tau}$ and the corresponding ground truth $\mathbf{{Y}^{1:\tau}}$.

\section{Experiments} \label{sec:exp}

We then conduct a series of experiments on \texttt{SINPA} to validate the efficacy of DeepPA. These experiments are designed to address the following Research Questions (RQ):
\begin{itemize}[leftmargin=*]
    \item $\textbf{RQ1}$: How does DeepPA compare in performance to existing PA forecasting approaches?
    \item $\textbf{RQ2}$: How does each module within DeepPA contribute to improving overall model performance?
    \item $\textbf{RQ3}$: What efficiency gains are achieved through the implementation of the GCO module?
    \item $\textbf{RQ4}$: Can DeepPA be effectively applied in practical scenarios for real-time online prediction?
\end{itemize}

\begin{table*}[!t]
  \centering
  \small
  \tabcolsep=0.5mm
  
    \begin{tabular}{l||c|c|cc|cc|cc|cc}
    \shline
    \multirow{2}*{\textbf{Model}} & \multicolumn{1}{c|}{\multirow{2}*{\textbf{\#Param}}} & \multicolumn{1}{c|}{\multirow{1}*{\textbf{Time /}}} & \multicolumn{2}{c|}{\textbf{Average (1$\sim$12 steps)}} & \multicolumn{2}{c|}{\textbf{0-1h (1$\sim$4 steps)}} & \multicolumn{2}{c|}{\textbf{1-2h (5$\sim$8 steps)}} & \multicolumn{2}{c}{\textbf{2-3h (9$\sim$12 steps)}} \\
\cline{4-11}    &      &  \textbf{epo (s) $\downarrow$}  & \textbf{MAE $\downarrow$} & \textbf{RMSE $\downarrow$} & \textbf{MAE $\downarrow$} & \textbf{RMSE $\downarrow$} & \textbf{MAE $\downarrow$} & \textbf{RMSE $\downarrow$} & \textbf{MAE $\downarrow$} & \textbf{RMSE $\downarrow$} \\
    \hline
    \hline
    HA & -  & - & 45.28 & 69.68  & 44.41 &  68.11 & 43.68 & 67.89 & 45.62 & 71.44  \\
    VAR & -  & - & 54.12 & 105.64  & 53.10 & 93.98 & 56.86 & 94.46 & 55.71 & 130.00  \\
    \hline
    DCRNN & 392 & 896 & 12.68{\tiny $\pm$0.02} & 30.33{\tiny $\pm$0.05} & 9.53{\tiny $\pm$0.04} & \underline{27.16}{\tiny $\pm$0.06} & 12.80{\tiny $\pm$0.04} & \underline{30.04}{\tiny $\pm$0.09} & 15.71{\tiny $\pm$0.03} & 33.83{\tiny $\pm$0.06}    \\
    STGCN & 583 & \underline{221} & 12.70{\tiny $\pm$0.11} & 31.17{\tiny $\pm$0.20} & 9.64{\tiny $\pm$0.06} & 27.71{\tiny $\pm$0.09} & 12.94{\tiny $\pm$0.18} & 31.03{\tiny $\pm$0.23} & 15.65{\tiny $\pm$0.23} & 34.84{\tiny $\pm$0.31}   \\
    GWNET & 808 & 303 & 12.76{\tiny $\pm$0.21} & 31.14{\tiny $\pm$0.18} & 9.54{\tiny $\pm$0.14} & 27.73{\tiny $\pm$0.15} & 12.91{\tiny $\pm$0.27} & 30.88{\tiny $\pm$0.26} & 15.84{\tiny $\pm$0.28} & 34.86{\tiny $\pm$0.16} \\
    MTGNN & 207 & 361  & 12.41{\tiny $\pm$0.24} & 30.62{\tiny $\pm$0.18} & 9.31{\tiny $\pm$0.16} & 27.37{\tiny $\pm$0.10} & 12.50{\tiny $\pm$0.25} & 30.37{\tiny $\pm$0.25} & 15.42{\tiny $\pm$0.34} & 34.16{\tiny $\pm$0.33}   \\
    ASTGCN & 11,587 & 540  & \underline{12.06}{\tiny $\pm$0.23} & \underline{30.22}{\tiny $\pm$0.31} & 9.88{\tiny $\pm$0.12} & 28.37{\tiny $\pm$0.19} & \underline{12.27}{\tiny $\pm$0.24} & 30.21{\tiny $\pm$0.32} & \underline{14.02}{\tiny $\pm$0.34} & \underline{32.10}{\tiny $\pm$0.41}   \\
    \hline
    Du-Parking & 392 & 683  & 12.55{\tiny $\pm$0.04} & 30.41{\tiny $\pm$0.05} & \underline{9.22}{\tiny $\pm$0.02} & \textbf{27.09}{\tiny $\pm$0.07} & 12.62{\tiny $\pm$0.04} & 30.12{\tiny $\pm$0.08} & 15.80{\tiny $\pm$0.07} & 34.05{\tiny $\pm$0.08}   \\
    SHARE &  476 & 402  & 12.57{\tiny $\pm$0.05} & 31.02{\tiny $\pm$0.21} & 9.33{\tiny $\pm$0.04} & 27.46{\tiny $\pm$0.03} & 12.70{\tiny $\pm$0.05} & 30.74{\tiny $\pm$0.17} & 15.69{\tiny $\pm$0.07} & 34.90{\tiny $\pm$0.43}   \\
    \hline
    \rowcolor{background_gray}
        DeepPA (ours) & 197 & \textbf{219}  & \textbf{10.95}*{\tiny $\pm$0.10} & \textbf{29.54}*{\tiny $\pm$0.06} & \textbf{9.11}*{\tiny $\pm$0.09} & 28.00{\tiny $\pm$0.07} & \textbf{11.18}*{\tiny $\pm$0.09} & \textbf{29.57}*{\tiny $\pm$0.08} & \textbf{12.56}*{\tiny $\pm$0.15} & \textbf{31.06}*{\tiny $\pm$0.11}   \\
    \shline
    \end{tabular}%
    \caption{5-run results. The \textbf{bold}/\underline{underlined} font means the best/the second best result. \#Param: the number of parameters, expressed in thousands (Kilo). * denotes the improvement over the second best model is statistically significant at level 0.05 \protect.}
  \label{tab:results}%
\end{table*}%

\subsection{Dataset Description}

We crawled over three-year real-time PA data (2018/03/26 to 2021/08/11) every 5 minutes from 1,921 parking lots throughout Singapore from Data.gov.sg\footnote{https://data.gov.sg/}. For our experiments, we re-sampled the raw dataset into the 15-minute interval and chose lots with a missing rate of PA of less than 30\%. In addition, due to the temporal distribution shift, we only use one-year data (2020/07/01 to 2021/06/30), and the ratio of training: validation: testing sets is set as 10:1:1. We then remove parking lots with obvious distribution shift (i.e., high KL divergence). After sample filtering, it remains 1,687 parking lots with stationary data distributions. We also crawl external attributes for these lots, including meteorological data (i.e., temperature, humidity, and wind speed), panning areas, utilization type, and road networks data from Data.gov.sg, the Urban Redevelopment Authority (URA)\footnote{https://www.ura.gov.sg/} and the Land Transport Authority (LTA) website\footnote{https://datamall.lta.gov.sg/content/datamall/en.html}.

\subsection{Experimental Settings}
We implement our model by PyTorch 1.10 using a Quadro RTX A6000 GPU. The Adam optimizer is utilized to train our model, and the batch size is 8. The learning rate starts from $1\times10^{-3}$, halved every three epochs. For the hidden dimension $C$ in SLBlock and SLBlock, we conduct a grid search over $\left\{8,16,32,64,128\right\}$, and $C$ = 64 obtains the best result. The block number of SLBlock and TLBlock is 2.

We compare our DeepPA with the following baselines that belong to the following four categories: (1) \emph{Classical methods}: \textbf{HA} \cite{zhang2017deep} and \textbf{VAR} \cite{toda1991vector}. (2) \emph{STGNN variants}: \textbf{DCRNN} \cite{li2018dcrnn_traffic}, \textbf{STGCN} \cite{yu2018spatio}, \textbf{GWNET} \cite{wu2019graph}, and \textbf{MTGNN} \cite{wu2020connecting}. (3) \emph{Attention-based models}: \textbf{ASTGCN} \cite{guo2019attention}. (4) \emph{Deep learning for PA prediction}: \textbf{SHARE} \cite{zhang2020semi} and \textbf{Du-parking} \cite{rong2018parking}. 

Note that some strong baselines \textbf{STTN}\cite{xu2020spatial} \textbf{GMAN}\cite{zheng2020gman}, are omitted due to Out-Of-GPU-Memory (OOM) caused by a large number of carparks within \texttt{SINPA}. 

\subsection{Model Comparison (RQ1)}

To address $\textbf{RQ1}$, following previous studies, we perform a model comparison in terms of Mean Absolute Error (MAE) and Root Mean Squared Error (RMSE)\cite{jean2016combining}. We run each method five times and report the average metric of each model. As shown in Table~\ref{tab:results}, DeepPA significantly outperforms all competing baselines on both metrics according to the Student’s T-test at level 0.05. 

In terms of performance, we can observe that: 1) Deep-learning-based approaches show a substantial improvement over classical methods like HA and VAR due to their enhanced learning capacity. 2) Our proposed model exhibits a pronounced advantage over the SHARE and Du-parking models in the domain of parking space prediction, thereby providing additional validation for the influential role of the collected spatio-temporal information on PA prediction. 3) In comparison to the second-ranked approach (i.e., ASTGCN), DeepPA demonstrates a remarkable 9.2\% decrease in MAE for long-term future prediction within the time interval of 3h. 4) Shifting the focus to individual hour predictions, the performance improvements for a single hour are observed to be 0.1\%, 8.8\%, and 10.4\%, respectively.

Moreover, we have evaluated the efficiency of each method by reporting both the number of parameters and the training time per epoch, as detailed in Table~\ref{tab:results}. Our proposed model stands out due to its minimal parameter count and the fastest training duration among all baselines. This underscores our model's suitability for practical deployment: the reduced parameter requirement minimizes memory demands, making it ideal for resource-constrained environments;  the rapid training capability highlights its ability to quickly adapt to user needs, significantly enhancing user experience. 

These attributes collectively make our model an optimal choice for real-world applications where \textit{accuracy}, \textit{efficiency}, and \textit{adaptability} are of paramount importance.

\subsection{Ablation Study (RQ2)}

\subsubsection{Effects of SLBlock}

\begin{figure}[t]
      \centering
      \includegraphics[width=0.9\linewidth]{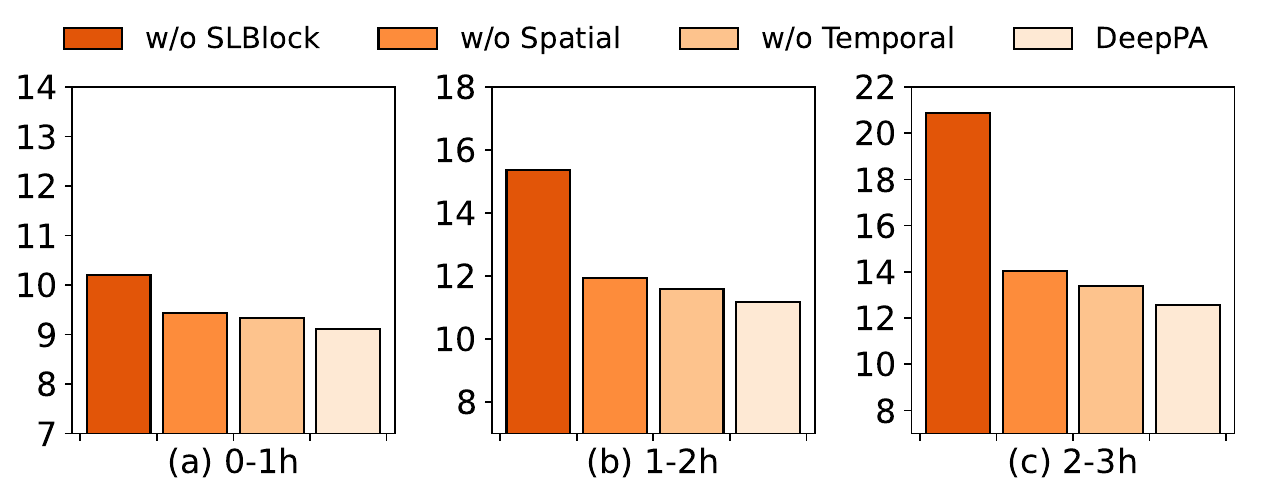}
      \caption{Effects of SLBlock on MAE.}\label{fig:ablation_s}
    \end{figure}

To investigate the effects of SLBlock, we consider the following variables for comparison: a) \textbf{w/o SLBlock}: we replace SLBlock with regular convolution solely operating on the PA. b) \textbf{w/o spatial}: we do not treat spatial information as a distinct entity for interaction with the PA. c) \textbf{w/o temporal}: we do not incorporate time information as a separate node for interaction with PA. The results (see Figure~\ref{fig:ablation_s}) reveal that removing either SLBlock or the ST information within the module leads to a significant decrease in MAE, underscoring the critical role of modeling spatial dependencies.

\subsubsection{Effects of TLBlock}
\begin{figure}[t]
      \centering
      \includegraphics[width=0.9\linewidth]{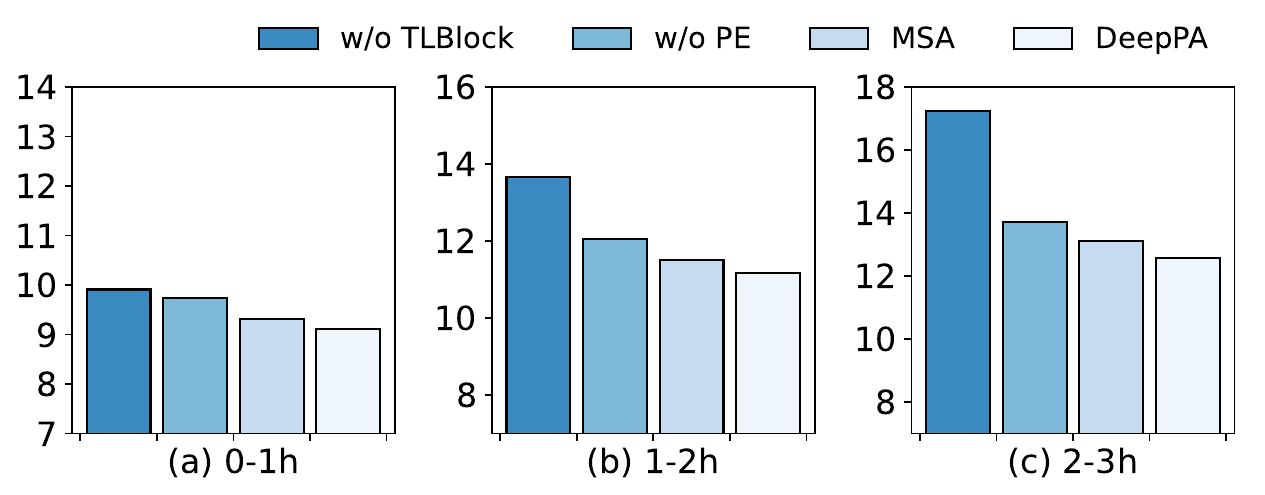}
      \caption{Effects of TLBlock on MAE.}\label{fig:ablation_t}
    \end{figure}

We then compare our model with its variants integrated with various temporal modules to assess the effectiveness of our TLBlock as well as the components in it: a) \textbf{w/o TLBlock}: we replace TLBlock with regular convolution solely operating on the PA. b) \textbf{MSA}: we replace Causal MSA with vanilla MSA. c) \textbf{w/o PE}: we remove the temporal position encoding from TLBlock. The results (see Figure~\ref{fig:ablation_t}) show that all the variants incorporating temporal modules demonstrate superior performance compared to the variants without TLBlock, confirming the importance of temporal information. 

\subsubsection{Hyperparameter Study}

\begin{figure}[t]
  \centering
  \includegraphics[width=0.85\linewidth]{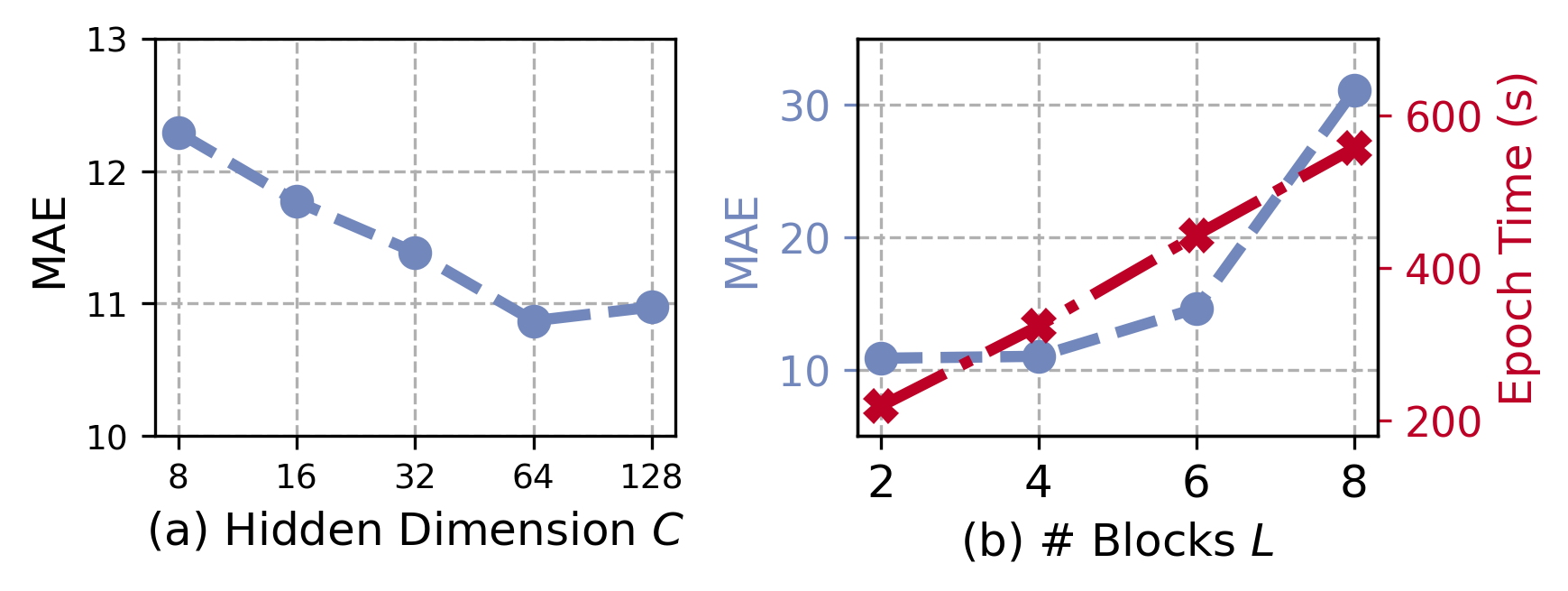}
  \caption{Effects of (a) hidden dimension $C$ and (b) the number of blocks $L$.}\label{fig:hyper}
\end{figure}

We explore the effects of the number of DeepPA blocks $L$ and the hidden dimension $C$. First, we fix the block size at 2 and increase the hidden dimension starting from 8. Results in Figure \ref{fig:hyper}a show that once the hidden dimension reached 64, the performance did not improve further. We thus set $C$ as 64 and increase $L$ starting from 2 (see Figure \ref{fig:hyper}b) and found that when $L$ increases from 2 to 4, there is no significant improvement in the performance. However, the training time per epoch increases by nearly double. Furthermore, when $L$ exceeds 4, the model's performance deteriorates sharply, possibly due to overfitting and other factors.

\subsection{Efficiency Study (RQ3)}
To address $\textbf{RQ3}$, we evaluate the efficiency gains achieved by the GCO compared with other methods, then examine how the performance varies with the retention of different low-frequency signals.

\subsubsection{Efficiency of GCO}
To investigate the efficiency improvements introduced by the GCO, we replace this module with the following modules: a) \textbf{MSA}: while effective in capturing relationships across all parking lots, MSA is known for its high theoretical time complexity. b) \textbf{AFNO}: conceptually similar to GCO, AFNO \cite{guibas2021efficient} uses neural operators to approximate MSA. Previous research has shown its effectiveness in addressing the high time complexity issues in Vision Transformers with increased image resolution. Table~\ref{tab:GCO} shows the training time per epoch for each variant. Compared to MSA and AFNO, GCO not only maintains performance but also significantly improves efficiency. Notably, after removing the SLBlock, the model's training time per epoch is around 115s, indicating that GCO reduces the training time by about 3.20 times and 2.37 times compared to MSA and AFNO, respectively, demonstrating GCO's efficient capability. 

\begin{table}[t]
  \centering
  \small
  \tabcolsep=1.2mm
  \caption{Comparative efficiency of GCO against other methods, measured in MAE. \#Param: the number of parameters, expressed in thousands (Kilo).}
    \begin{tabular}{l||c|c|c|c|c|c}
    \shline
     \multicolumn{1}{c||}{\multirow{1}*{\textbf{Variant}}} & \multicolumn{1}{c|}{\multirow{1}*{\textbf{1-3h}}} & \multicolumn{1}{c|}{\textbf{0-1h}} & \multicolumn{1}{c|}{\textbf{1-2h}} & \multicolumn{1}{c|}{\textbf{2-3h}} & \multicolumn{1}{c|}{\textbf{\#Param}} & \multicolumn{1}{c}{\textbf{Time(s)}} \\
    \hline
    \hline
      MSA & \textbf{10.87} & \textbf{9.07} & \textbf{11.10} & \textbf{12.44} & 227 & 448 \\
      AFNO & 10.90 & 9.11 & 11.12 & 12.48 & 199 & 361 \\
      \rowcolor{background_gray}
      GCO & 10.95 & 9.11 & 11.18 & 12.56 & 197 & \textbf{219} \\
    \shline
    \end{tabular}%
  \label{tab:GCO}%
\end{table}%

\subsection{Practicality (RQ4)}

\begin{figure}[t]
  \centering
  \includegraphics[width=\linewidth]{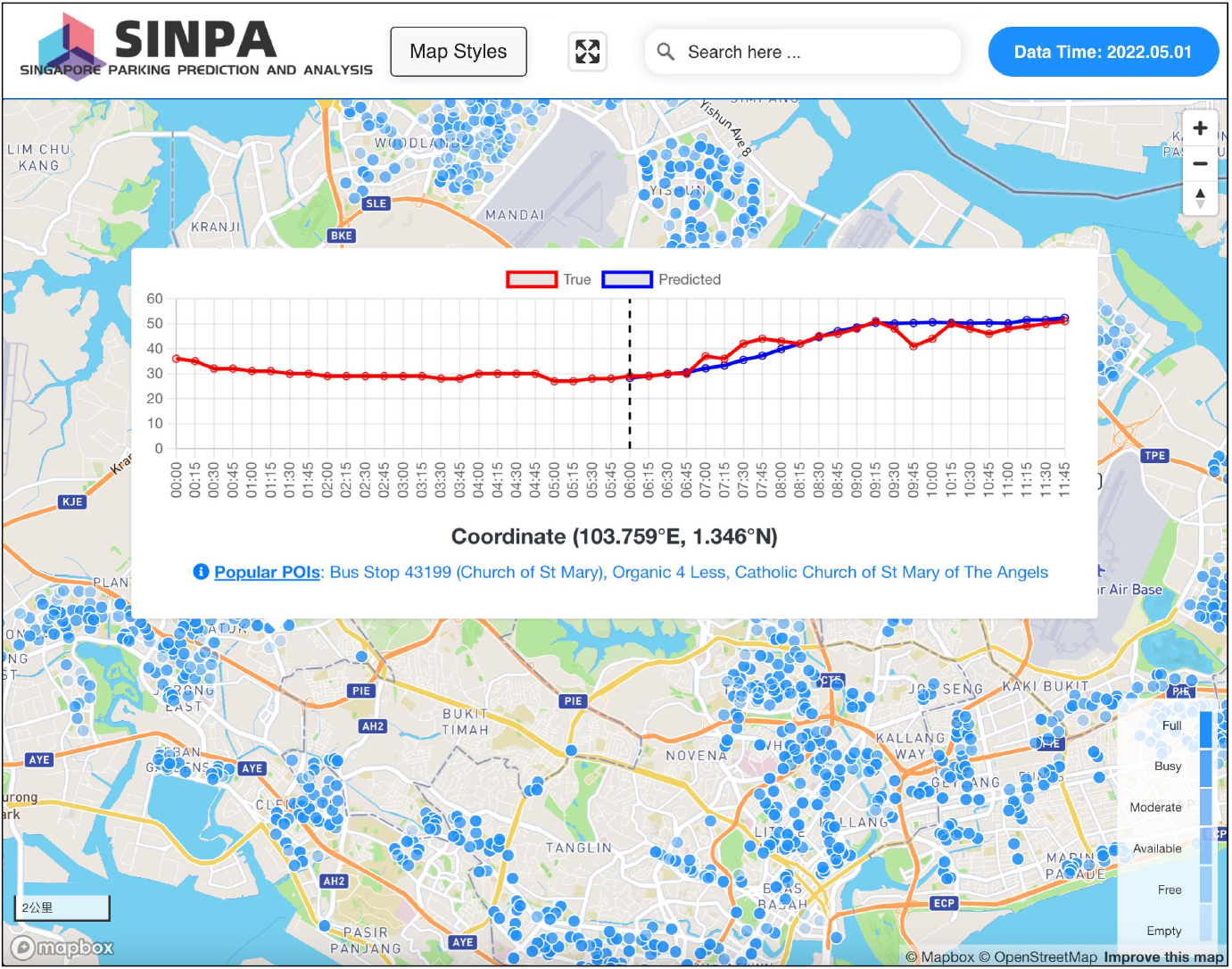}
  \caption{Our web-based PA monitoring and prediction system.}\label{fig:sinpa_system}
\end{figure}

Regarding $\textbf{RQ4}$, leveraging the demonstrated efficiency and accuracy of our model, we deploy it for real-time parking availability forecasting. 
An illustrative example of our system in action is shown in Figure~\ref{fig:sinpa_system}. The web is built on the Mapbox platform \cite{rzeszewski2023mapbox}. After selecting the target parking space, the system will predict future PA and present it in the form of a line graph. The shown graph displays a notable correlation between the predicted (represented by the blue line) and actual (represented by the red line) PA readings, demonstrating our model's high accuracy in predicting PA trends and variations effectively. For a direct experience of this interactive forecasting tool, please visit our demo web-based platform at \url{https://sinpa.netlify.app}.

\section{Conclusion and Future Work}

In this paper, we crawled, processed, and released the \texttt{SINPA} dataset, a comprehensive collection of one-year PA readings from over 1,600 parking lots across Singapore, along with various external factors. To the best of our knowledge, \texttt{SINPA} is the first dataset of its kind to be made publicly available. Considering the complex relationships and the important role of external features, we presented DeepPA, a deep-learning model tailored for PA prediction. Our evaluations demonstrate the accuracy, efficiency, and adaptability of our model. In the future, we plan to explore reinforcement learning to improve parking recommendation services.

\section*{Acknowledgement}
This work is mainly supported by a grant from State Key Laboratory of Resources and Environmental Information System and the Guangzhou-HKUST(GZ) Joint Funding Program
(No. 2024A03J0620). This work is also funded by the Advanced Research and Technology Innovation Centre (ARTIC), the National University of Singapore under Grant (project number: A-8000969-00-00).
\bibliographystyle{named}
\bibliography{ijcai24}

\end{document}